\documentclass{article} 
\usepackage{iclr2027_conference,times}

\usepackage{amsmath,amsfonts,bm}

\def\eqref#1{equation~\ref{#1}}

\def\1{\bm{1}}

\DeclareMathAlphabet{\mathsfit}{\encodingdefault}{\sfdefault}{m}{sl}
\SetMathAlphabet{\mathsfit}{bold}{\encodingdefault}{\sfdefault}{bx}{n}

\usepackage{siunitx} 
\usepackage{xspace}
\usepackage{booktabs}
\usepackage{pifont}
\usepackage{url}
\usepackage{multirow}
\usepackage{graphicx}
\usepackage{float}
\usepackage{placeins}
\usepackage{tikz}
\usetikzlibrary{positioning, arrows.meta, fit, calc}
\usepackage{mathtools}
\usepackage{hyperref}
\usepackage{needspace}

\newcommand{\boldsub}[1]{\noindent\textbf{#1}\ }

\newcommand{\answerTODO}[1][]{\textcolor{red}{\bf [TODO]}}
\newcommand{\justificationTODO}[1][]{\textcolor{red}{\bf [TODO]}}

\title{ On Hyperparameter Tuning on the Test Set  }

\author{Matteo Fregonara, Tom Viering, Jan van Gemert
\\
Delft University of Technology\\
}

\iclrfinalcopy 
\begin{document}

\maketitle

\begin{abstract}
"\textit{Don't tune hyperparameters on the test set}" is often stated in machine learning textbooks. Violating it is considered a cardinal sin that produces misleadingly optimistic results, corrupts benchmark integrity, and thus can even be interpreted as scientific fraud. Yet evidence suggests that test set hyperparameter tuning does occur in practice, making it all the more important to understand its actual consequences. So how bad is it, really? In this work we question this dogma and put it to an empirical test. We systematically study the magnitude of the performance inflation caused by tuning the hyperparameters on the test set for MNIST-1D, CIFAR-10, and three tasks from the GLUE benchmark. Our experiments show that while the effect is real and significant, it is frequently small relative to other sources of noise. In many cases, we find that tuning all models on the test set recovers exactly the same model as when tuning all models on the validation set. Most importantly, we find that the rankings of models remain essentially preserved after tuning on the test set and therefore that consistent test-set tuning may not invalidate benchmarks or model selection. Our results call for a more nuanced view of tuning hyperparameters on the test set. We do not aim to normalize test-tuning because it adds a clear bias, but we do hope that our results stimulate researchers  to openly report that they do test tuning if it does occur.
\end{abstract}

\section{Introduction}

Overfitting is an inflated generalization estimate, possibly distorting  model selection rankings, which may lead to deploying the wrong model.  We investigate overfitting of hyperparameters:  configurations that determine model training behavior. Specifically, we investigate overfitting by tuning all hyperparameters on the test set, \emph{i.e.}, the same data set used for reporting  generalization performance.

Machine learning textbooks prohibit hyperparameter tuning on the test set. From a theoretical perspective, a hyperparameter is no different than any other model parameter that is set at training time. Thus, an extreme view can see tuning hyperparameters on the test set as academic fraud in the current benchmarking paradigm~\citep{hardt2025mlBenchmarkScience}. E.g, tuning parameter initialization on the test set~\citep{picard2021seed} gives excellent accuracy; independent of other claims for empirical gain~\citep{lipton2019research, sculley2018winnerCurse}.

In practice, several implementations already tune hyperparameters on the test set, both in computer vision~\citep{pytorch_imagenet_example, timm_repo, deit_repo, beit_repo, mae_repo} and in natural language processing~\citep{devlin2019bert, google_bert_repo, liu2019roberta, fairseq_roberta_glue}. Thus, it can be argued that a hyperparameter is \textit{not} like a normal model parameter, because they usually effect a model globally, and only a handful hyperparameters configurations are evaluated~\citep{hastie2009elements}. Thus, there is tension between theory and practice, and here we investigate how bad hyperparameter overfitting really is.

In this paper, we give the first systematic empirical evaluation of overfitting by tuning all hyperparameters on the test set. We carefully construct a controlled evaluation benchmark of repeated hyperparameter tuning on a validation set, compared to tuning on the test set. We evaluate on the controlled MNIST-1D~\citep{greydanus2024scaling}, image classification  on CIFAR-10~\citep{krizhevsky2009learning} and three natural language understanding tasks from the GLUE benchmark~\citep{wang2019glue} (RTE, MRPC and CoLA). We test multiple model families, and train $36{,}000$ model variants. We investigate how easily models overfit, how severe the overfitting is, and if the overfitting leads to differences in model selection.

Our results show that overfitting by tuning on the test set is significant on every benchmark we evaluate. Yet, the accuracy gain by overfitting is mild when compared to other sources of randomness such as data splits. Model selection rankings barely change when all models are tuned on test data, and we did not observe that the best model is not selected.

Our results call for a nuanced view on tuning hyperparameters on the test set. Overfitting is real, and comparing validation-tuned models vs test-tuned models is unfair, as it clearly benefits test-tuned models. However,  model rankings rarely change when \textit{all} models under comparison are tuned on test data, which makes the effect of test-tuning relatively benign when comparing test-tuned models to other test-tuned models. Our results benefit the community in the following two ways: (1) we validate that existing test-tuned results are not inherently flawed, and have value, at least when compared to other test-tuned results. (2) If new research does validation-tuning and experimentally needs to compare to published test-tuned work, then this comparison is unfair and publishing becomes difficult. Our paper shows that if authors openly mention they test-tune, that comparing test-tuned models to each other is reasonable, offering a way out for new papers to compare to published test-tuned papers.

\section{Related Work}

\paragraph{Adaptive overfitting and test set reuse: theory vs practice.} 

Theory suggests~\citep{dwork2015generalization, makarova2021overfitting}  that long-term reuse of benchmark test sets should lead to measurable overfitting. In practice, however, empirical evidence paints a more nuanced picture. New test sets for CIFAR-10 and ImageNet, constructed to match the original data collection process, show that accuracy drops of 3--15\% on CIFAR-10 and 11--14\% on ImageNet are largely explained by a distribution shift instead of adaptive overfitting~\citep{recht2018cifar, recht2019imagenet_generalize}. Importantly, model rankings are preserved, with an approximately linear relationship between original and replicated accuracy.
Evidence at a larger scale points in the same direction~\citep{liao2021learning_eval_failures,roelofs2019meta_overfitting}. Selecting a model on a holdout set and reporting on that same set is known to inflate the estimate, both under cross-validation~\citep{varma2006bias, JMLR:v11:cawley10a} and under a fixed split~\citep{nagler2024reshufflingresamplingsplitsimprove}. Recent work has begun to formalize this risk for hyperparameter tuning~\citep{schroder2025overfitting, schneider2025overtuning, makarova2021overfitting}, and \citet{koleiev_end-user_2026} show that a single deliberately test-tuned model climbs into the top ranks of the ADMET leaderboards. No study, however, has isolated the effect when every model under comparison is tuned on the test set. We follow these works in a stronger setting: we give the tuning procedure direct access to the test set, and evaluate the theoretical claims, empirically.



\textbf{Test set hyperparameter tuning is common practice.} In practice, tuning and final reporting  is performed on the same labels, because official test labels are often withheld: ImageNet uses the \texttt{val} split as a proxy test set~\citep{russakovsky2015imagenet}, and GLUE~\citep{wang2019glue} the development split. The official PyTorch example~\citep{pytorch_imagenet_example} evaluates on \texttt{val} every epoch and keeps the best checkpoint, and others present hyperparameter search as an ablation on that same data~\citep{timm_repo, convnext_repo}. The DeiT~\citep{deit_repo, touvron2021deit}, BEiT~\citep{beit_repo, bao2022beit} and MAE~\citep{mae_repo, he2022mae} implementations go further, calling \texttt{val} the ``test'' set in their own print statements and metric keys. On GLUE, BERT~\citep{devlin2019bert, google_bert_repo} performs an exhaustive search choosing ``the model that performs best on the development set'' and, for the smaller tasks, ``several random restarts'' selecting ``the best model on the Dev set'', while RoBERTa~\citep{liu2019roberta, fairseq_roberta_glue} keeps only the checkpoint that maximises validation accuracy and reports on that same split. We hypothesize that these works tune on the test set because the computational demand of current models have become so large, that they assume that comparing test-tuned models to other test-tuned models is reasonable. This is the hypothesis that our paper empirically explores. Are these test-tuned results flawed, or do they still have value?

\boldsub{Hyperparameter sensitivity.} 
Sensitivity analysis methods quantify how much hyperparameter tuning can cause overfitting. For example learning rate decay and batch size~\citet{sadl_sensitivity}, or the number of convolutional layers, learning rate, dropout rate, optimizer and number of epochs~\citep{wang2024hyperparameter_importance}, and their interactions~\citep{sha2019hyperparameter}. Thus, in our experimental setup we explictly validate that the hyperparameters we include in the study are sensitive to overfitting, i.e., we first make sure if they sufficiently allow for overfitting when unfairly comparing validation-tuning to test-tuning. Then, we evaluate overfitting when comparing test-tuning to test-tuning.

\textbf{Hyperparameter tuning leads to significant improvements.} Empirical evidence shows that hyperparameter tuning is a major driver of model performance~\citep{weerts2020importance}. For example, tuning only learning rate, weight decay and batch size on a validation split results in a baseline that outperforms several specialized methods~\citep{brigato2021tune}, which suggests that reported gains may partly reflect more thorough hyperparameter optimization rather than algorithmic innovation~\citep{lipton2019research, sculley2018winnerCurse}. The process is also heavily shaped by human skill~\citep{anand2020blackMagicinDL}. 
The importance of hyperparameter tuning motivates our work.

\boldsub{Variance in benchmarks.} 
Benchmark accuracy varies due to data sampling, model initialization, training dynamics and hyperparameter optimization. If this variability is not accounted for, comparisons between models can be biased because observed differences may reflect noise rather than systematic effects~\citep{bouthillier2021variance}.
To address this, models can be compared using the probability of outperforming. Let $A$ and $B$ be the performances of two models to be compared. Then $P(A > B)$ can be estimated over repeated runs, where all sources of randomness should be varied. This provides a more robust measure than single-run accuracy and allows uncertainty to be quantified explicitly. This is the setup that we use in our paper.

Empirical evidence shows that  variability can be substantial. A sweep over up to $10^4$ random seeds on CIFAR-10 and ImageNet~\citep{picard2021seed} finds that, despite low average variance, it is easy to obtain outlier seeds with significantly better or worse accuracy. Differences of performance in literature can therefore arise purely from randomness or overfitting. Here, we question if these benchmark rankings are still valid when all hyperparameters are tuned on the test set.

\section{Methodology}
\label{sec:methodology}


\paragraph{Theoretical motivation.} There are no theoretical guarantees that selecting the best model by hyperparameter tuning on a validation set also selects the best model for the test set~\citep{bengioJMLR04noUnbiased}. In practice, however, there is the common assumption that they do. Here we make these model selection assumptions explicit, and motivate our empirical research. 

We focus on a classification setting and let $D_{\text{train}}$, $D_{\text{val}}$ and $D_{\text{test}}$ denote training, validation and test datasets, respectively, drawn \emph{i.i.d} from the unknown distribution $\mathcal{D}$. The hyperparameter space for $N$ hyperparameters is $\Lambda$, where a configuration $\lambda \in \Lambda$ is an $N$-dimensional vector, and $A(\lambda, D)$ represents the accuracy of a learning algorithm $A$ trained on $D_{\text{train}}$ with hyperparameters $\lambda$ and evaluated on dataset $D$. 

Hyperparameter tuning for model selection is about making comparisons between hyperparameters and choosing the best. In the theoretical case, when we assume $D_{\text{val}} = D_{\text{test}}= \mathcal{D}$, and $A$ has no stochasticity, then the probability of making a comparison between pairs of hyperparameters $\{\lambda_a, \lambda_b\} \in \Lambda$ is identical between $D_{\text{val}}$ and $D_{\text{test}}$, 
\begin{equation}
\label{eq:infinite}
P_{\substack{\{\lambda_a, \lambda_b\} \in \Lambda}} \big( A( \lambda_a, D_{\text{val}} )   \ge A( \lambda_b, D_{\text{val}} ) \big) = P_{\substack{\{\lambda_a, \lambda_b\} \in \Lambda}}  \big( A( \lambda_a, D_{\text{test}})   \ge A( \lambda_b, D_{\text{test}}) \big).
\end{equation}
Because comparisons on the validation set are the same as comparisons on the test set, this allows optimizing accuracy on the validation set to optimize accuracy on the test set.

In practice, datasets are finite, and $D_{\text{val}} \neq D_{\text{test}}$, which will introduce accuracy differences due to accidental dataset sampling differences. To take these into account, we assume $D_{\text{val}}$ and $D_{\text{test}}$ have the same number of samples, and decompose the accuracy for hyperparameters $\lambda$  in a term for the algorithm  $A( \lambda, \mathcal{D} )$ and terms for the accidental dataset bias $B( \lambda, D_{\text{val}} )$ and $B( \lambda, D_{\text{test}} )$. Then, the practical assumption that allows tuning hyperparameters $\lambda_a$ and $\lambda_b$  on the validation set is 
\begin{align}
\label{eq:finite}
P_{\substack{\{\lambda_a, \lambda_b\} \in \Lambda}}  \big( A( \lambda_a, \mathcal{D} ) + B( \lambda_a, D_{\text{val}} )   & \ge  A( \lambda_b, \mathcal{D} ) + B( \lambda_b, D_{\text{val}} ) \big)  \approx \\
\label{eq:testsetassume}
P_{\substack{\{\lambda_a, \lambda_b\} \in \Lambda}}  \big( A( \lambda_a, \mathcal{D} ) + B( \lambda_a, D_{\text{test}})   & \ge  A( \lambda_b, \mathcal{D} ) + B( \lambda_b, D_{\text{test}}) \big).
\end{align}

This assumes that algorithmic differences  in the same dataset are larger than dataset bias differences, 
\begin{align}
\label{eq:assume}
|A( \lambda_a, \mathcal{D} ) - A( \lambda_b, \mathcal{D} )| & > \max \big( |B( \lambda_a, D_{\text{val}} ) - B( \lambda_b, D_{\text{val}} )|, |B( \lambda_a, D_{\text{test}} ) - B( \lambda_b, D_{\text{test}} )| \big) 
\end{align}
because otherwise biases due to accidental data sampling dominate model selection.  This assumption allows finding the best hyperparameters on the validation set, which then approximately will give the best hyperparameters on the test set. 

Model selection on the test set can take two forms. One form is comparing one model on the test set, with other models from the validation set, as 
\begin{equation}
A( \lambda_a, \mathcal{D} ) + B( \lambda_a, D_{\text{test}} )    >  A( \lambda_b, \mathcal{D} ) + B( \lambda_b, D_{\text{val}} ).
\end{equation}
This is the form investigated most in existing literature, e.g.,~\citep{schroder2025overfitting, schneider2025overtuning, makarova2021overfitting} and is unfair, because it allows to exploit accidental dataset sampling biases between sampled datasets by optimizing for $ |B( \lambda_a, D_{\text{test}} ) - B( \lambda_b, D_{\text{val}} )| $. 

The other form of test tuning is what we investigate here:
\begin{equation}
A( \lambda_a, \mathcal{D} ) + B( \lambda_a, D_{\text{test}})    >  A( \lambda_b, \mathcal{D} ) + B( \lambda_b, D_{\text{test}}).
\end{equation}
Where dataset biases are restricted to be within the same dataset, and is not breaking the assumption in Eq.~\ref{eq:assume} and is part of the assumption in Eq.\ref{eq:testsetassume}. Since there are no theoretical guarantees, we empirically investigate  the severity of this type of test set tuning and how it influences model selection, in practice.

\paragraph{Problem setup and selection strategies} We use the setting for which we have evidence that hyperparameters are tuned on the test set, both in computer vision~\citep{pytorch_imagenet_example, timm_repo, deit_repo, beit_repo, mae_repo} and in natural language processing~\citep{devlin2019bert, google_bert_repo, liu2019roberta, fairseq_roberta_glue}. These works use independent hyperparameter sampling, and thus we do too. Hyperparameters are selected by sampling $T$ candidate configurations $\Lambda_T = \{\lambda_1, \dots, \lambda_T\} \subset \Lambda$ via random sampling. For a learning algorithm's accuracy $A$ we compare two hyperparameter selection strategies: \emph{validation-tuning} selects $\lambda^*_{\text{val}} = \arg\max_{\lambda \in \Lambda_T} A(\lambda, D_{\text{val}})$, while \emph{test-tuning} selects $\lambda^*_{\text{test}} = \arg\max_{\lambda \in \Lambda_T} A(\lambda, D_{\text{test}})$, where we carefully pair the validation-tuned and the test-tuned runs so that the only source of variation is in the evaluation dataset $D_{\text{val}}$ vs $D_{\text{test}}$. We quantify  overfitting through the \emph{overfitting gap} between tuning on the test set, versus tuning on the validation set,
\begin{equation}
  \Delta_{\text{OG}} \;=\; A(\lambda^*_{\text{test}}, D_{\text{test}}) \;-\; A(\lambda^*_{\text{val}}, D_{\text{test}}),
\end{equation}
the excess test accuracy attributable to hyperparameter selection with direct knowledge of $D_{\text{test}}$. Because $\lambda^*_{\text{test}}$ maximises $A(\lambda, D_{\text{test}})$ over the same pool $\Lambda_T$ from which $\lambda^*_{\text{val}}$ is drawn, $\Delta_{\mathrm{AO}} \geq 0$ in every paired run, with equality when both rules pick a configuration that score identically on $D_{\text{test}}$.

\paragraph{Paired Run Design.} Each of the $K$ paired (val-tuned, test-tuned) training runs executes the following pipeline:
\begin{enumerate}
  \item sample an independent bootstrap split $(D_{\text{train}}, D_{\text{val}}, D_{\text{test}})$ (Appendix~\ref{app:data-splitting} describes the splitting scheme),
  \item draw $T$ independent hyperparameter configurations $\Lambda_T$ via random sampling,
  \item for each configuration $\lambda \in \Lambda_T$, train a model on $D_{\text{train}}$ (from random initialisation for the image benchmarks, and by fine-tuning a pretrained checkpoint for the GLUE tasks) and record $A(\lambda, D_{\text{val}})$ and $A(\lambda, D_{\text{test}})$,
  \item select $\lambda^*_{\text{val}}$ and $\lambda^*_{\text{test}}$ from the same pool of $T$ trained models and compute $\Delta_{\mathrm{OG}}$.
\end{enumerate}

Because of the paired run design, both selection strategies operate on the same bootstrap split, so any difference between $\lambda^*_{\text{val}}$ and $\lambda^*_{\text{test}}$ isolates the effect of the selection rule alone. We use $K = 30$ paired runs per model, the minimum needed for the Noether-bound paired-comparison guarantee~\citep{bouthillier2021variance,noether1987sample}.

Following \citet{bouthillier2021variance}, every experiment varies the major sources of randomness under our control (data splits, hyperparameter samples, initialisation seeds) so that the empirical distribution of $\Delta_{\mathrm{OG}}$ reflects the full pipeline variance rather than a single favorable configuration.

\section{Experiments}
\label{sec:experiments}

We investigate overfitting by tuning hyperparameters directly on the test set, i.e., using the same dataset for hyperparameter selection and accuracy estimation. We first validate that our experimental setting is suitable to measure model sensitivity to overfitting. We then investigate how severe the overfitting could be, and if tuning all models on the test set leads to different model selection compared to tuning all models on the validation set.

\paragraph{Datasets.} We evaluate on two image classification benchmarks and three natural language understanding tasks. The controlled benchmark is MNIST-1D~\citep{greydanus2024scaling}, a 1D-signal variant of MNIST designed for controlled experiments in machine learning. We use a 4,000-example source set bootstrapped per training run, and test a family of eight fully-connected networks spanning $1.3\,\text{k}$ to $2 {\times} 10^{5}$ parameters (Appendix~\ref{app:architectures}), trained with Adam for 8,000 steps. We optimize learning rate, learning-rate decay rate (set to zero with probability $0.3$), weight decay, and the random seed used to initialize the network weights.

For CIFAR-10 we combine the training and test sets into a single 60,000-example pool, from which each training run draws an independent bootstrap split. We use 8 architectures from the CIFAR-10 leaderboard (Appendix~\ref{app:architectures}), covering ResNets~\citep{he2016deep}, Shake-Shake~\citep{gastaldi2017shake}, VGG~\citep{simonyan2014very}, MobileNetV2~\citep{sandler2018mobilenetv2} and ResNet-9~\citep{page2019how}. We tune learning rate, weight decay, momentum, number of training steps and the random seed used to initialise the network weights, using Nesterov SGD~\citep{pmlr-v28-sutskever13} with cosine-annealed learning rate.

We additionally evaluate three classification datasets from the GLUE benchmark~\citep{wang2019glue}: RTE (textual entailment), MRPC (paraphrase detection) and CoLA (grammatical acceptability). The official GLUE test splits are unlabeled and withheld on the leaderboard server, so we combine each dataset's train and validation sets into their own single pool (RTE: $2{,}767$, MRPC: $4{,}076$ and CoLA: $9{,}594$ examples), from which each training run draws an independent bootstrap split. All three behave alike and all three are reported in the main text.

We fine-tune eight pretrained transformers spanning $11.8$M to $141.9$M parameters (Appendix~\ref{app:architectures}) with AdamW and a linear warmup-decay schedule, tuning the fine-tuning hyperparameters: learning rate, weight decay, warmup ratio, number of epochs and the initialization seed. We score all three by accuracy, so that one metric applies across all five benchmarks. Accuracy is GLUE's official metric for RTE and MRPC. For CoLA the official metric is Matthews correlation.

\paragraph{Experimental settings.} We use $K = 30$ paired training runs per architecture, where we re-sample the train, val, and test sets differently for each of the $K$ runs. The hyperparameter search budget per run is $T = 30$, which is in line with the number of configurations typically evaluated in deep learning architecture training~\citep{hastie2009elements}. We select the best accuracy over the $T$ randomly sampled different hyperparameter configurations. All randomness derives from a single master seed. Hyperparameter search spaces are in Appendix~\ref{app:hpo-spaces} and additional implementation details in Appendix~\ref{app:implementation-details}. In total this yields $8 \times 30 \times 30 = 7{,}200$ trained models per benchmark, or $36{,}000$ across the five benchmarks. The full set of $36{,}000$ trained models and $72{,}000$ validation/test set evaluations used approximately $8{,}000$ GPU-hours.

\subsection{Exp 1: validating if the experimental setup captures overfitting}

Before we  investigate the overfitting effect of tuning on the test set we first verify that our experimental setup is suitable to capture overfitting by hyperparameter tuning.

We first validate if overfitting is possible on the data itself and we compute an upper bound by training and tuning models directly on the test data. Every architecture nearly memorizes its own test fold, achieving exactly $100\%$ accuracy on both image benchmarks, and between $98.9\%$ and $100\%$ accuracy on the GLUE tasks, whose test folds are smaller (Appendix~\ref{app:upper-bound}). There is ample room to overfit.

\begin{figure}[t]
    \centering
    \includegraphics[width=\linewidth]{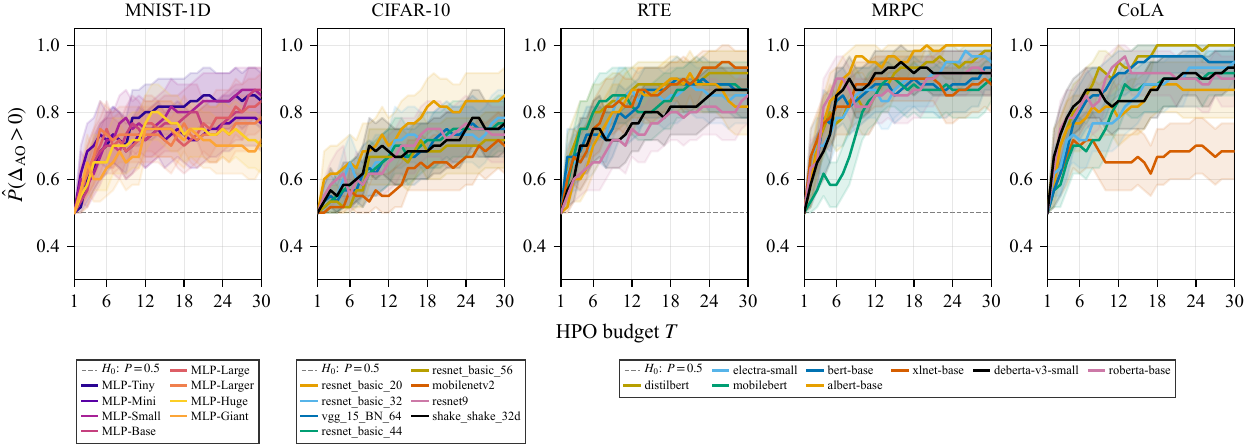}
    \caption{\textbf{Is the search budget $T$ sufficient to overfit through hyperparameter tuning?} The tie-broken win probability of test-tuning vs validation-tuning, $\hat{P}(\Delta_{\mathrm{AO}} > 0)$, as a function of the tuning budget $T$. Each line is one architecture, with its $95\,\%$ percentile-bootstrap CI shaded. A line whose CI lower bound clears the dashed threshold at $0.5$ shows overfitting at that budget. MNIST-1D and the three GLUE tasks clear it by $T{=}3$, CIFAR-10 at around $T{=}15$. All models overfit within the budget $T=30$. }
    \label{fig:p_vs_T_combined}
\end{figure}

\paragraph{Validating if the search budget is sufficient to overfit.} To evaluate if the hyperparameter search budget $T$ is large enough to overfit, we compare validation tuning to test tuning and evaluate $T$ against the empirical win probability of test-tuning over validation-tuning~\citep{bouthillier2021variance}, $\hat{P}(\Delta_{\text{OG}} > 0) \;\coloneqq\; P(\Delta_{\text{OG}} > 0) + \tfrac{1}{2}\, P(\Delta_{\text{OG}} = 0)$ where ties contribute $\frac{1}{2}$.  The rationale is that a model that overfits easily reaches a high win probability within our budget $T$.

Results in Figure~\ref{fig:p_vs_T_combined} show that all models easily overfit, needing at most $T{=}3$ different hyperparameter configurations on MNIST-1D and on all three GLUE tasks. For CIFAR-10, around $T{=}15$ is needed where also the overfitting probability is lower compared to the other datasets.  For all 5 datasets, all of the 40 models evaluated overfit within the budget. 

Note that $\hat{P}(\Delta_{\mathrm{AO}} > 0)$ is not strictly increasing with $T$, because a larger budget also benefits validation-tuning, which can in turn lower $\hat{P}(\Delta_{\mathrm{AO}} > 0)$. Because of its computational costs, we try a single model and reran MLP-Giant on MNIST-1D with $T = 100$ trials per run (a $3\times$ larger budget) and larger fixed splits of $10{,}000 / 10{,}000 / 10{,}000$ drawn from a $70{,}000$-sample MNIST-1D set, denoted MLP-Giant$_{T{=}100}$. Both findings hold at this budget: $\hat{P}(\Delta_{\mathrm{AO}} > 0) = 0.73$ at $T{=}100$ with the CI lower bound well above $0.5$. See also Appendix~\ref{app:robustness}.

Because CIFAR-10 seems to have a lower probability of overfitting, we re-evaluated the selected CIFAR-10 configurations on CIFAR-10.1~\citep{recht2019imagenet_generalize}, an independent $2{,}000$-image replication drawn from the same source distribution. There the val-tuned and test-tuned configurations score within run-to-run noise of each other, so the gap measured on the bootstrap fold does not carry over, while tuning directly on CIFAR-10.1 beats both by roughly $0.4\%$ (Appendix~\ref{app:cifar101}).

\paragraph{Validating that hyperparameter tuning can lead to changes in model rankings.} 

Figure~\ref{fig:p_vs_T_combined} show that all models easily overfit, but it does not show that the increased accuracy obtained by overfitting is enough to also lead to a change in model selection. It could be the case that model rankings are robust to hyperparameter tuning, and then we cannot conclude anything about test-tuning.

To evaluate ranking changes, we show ranking stability within individual runs in Table~\ref{tab:ranking-within-run} and final per-architecture rankings in Table~\ref{tab:rq1_per_arch}. Specifically, when we compare validation-tuned models to each other, in the \textit{val vs val} column, the hyperparameters allow for model reordering due to the randomness between each of the 30 runs. Furthermore, if we keep the validation randomness fixed and explicitly allow overfitting by comparing hyperparameter tuning on the validation set compared to tuning on the test set, in the \textit{val vs test} column, we show that unfair tuning leads to a change in the ranking for nearly half of the models: 18 out of 40. We see this in each dataset except CIFAR-10, which shows that the CIFAR-10 models are robust to hyperparameter tuning, and if CIFAR-10 models do not change rank when test-tuning, does not say much about the overfitting of test-tuning: it will not invalidate our results, but will not strengthen our results either. For the other 4 datasets, however, we validate that our experimental setting allows for significant model rank changes due to hyperparameter overfitting.

\begin{table}[t]
\centering
\begin{tabular}{lcccccc}
\toprule
 & \multicolumn{3}{c}{Avg pairs reordered (vs val-tuning)}
 & \multicolumn{3}{c}{Models overtaking (of 8)} \\
Dataset & val vs val & val vs test & test vs test & val vs val & val vs test & test vs test \\
\cmidrule(lr){1-1} \cmidrule(lr){2-4} \cmidrule(lr){5-7}
MNIST-1D & 5.07 & 3.40 & 2.10 & 0.41 & 3 & 0 \\
CIFAR-10 & 2.48 & 1.30 & 1.00 & 0.40 & 1 & 0 \\
RTE      & 5.18 & 4.50 & 2.50 & 0.80 & 5 & 0 \\
MRPC     & 7.21 & 8.63 & 3.60 & 0.76 & 5 & 1 \\
CoLA     & 3.98 & 3.73 & 2.00 & 0.34 & 4 & 1 \\
\bottomrule
\end{tabular}
\caption{\textbf{Ranking stability within individual runs.} In each of the $30$ runs we rank the eight architectures by validation-tuned accuracy and ask how that ordering changes under three comparisons. \textit{Val vs val}: both models keep their validation-tuned score, but the two rankings come from two different runs, so the column measures ordinary run-to-run variation with no test tuning at all. \textit{Val vs test}: one model is tuned on test while its opponent keeps only its validation-tuned score, which is unfair by construction and acts as a deliberate upper bound. \textit{Test vs test}: every model is tuned on test, so the comparison is fair, and this is the setting we study. \emph{Pairs reordered (left)} is computed per run, averaged over the $30$ runs and, for val vs val, over all $435$ pairs of runs. \emph{Models overtaking (right)} uses accuracy averaged over all runs and, for val vs val, is the mean over $10{,}000$ bootstrap resamples of the runs. We conclude the following. The capability gap of our CIFAR-10 models is too large to be effected by hyperparameter tuning. This does not invalidate our results, but CIFAR-10 models do not change rank, even when we try, and thus can also not be expected to do so when test-tuning. \textit{Val vs val} shows that there is sufficient random variation between each run. \textit{Val vs test} validates our experimental setup because it shows that explicit overfitting will lead to model rank changes, where for each dataset except CIFAR-10, around half of the models are overtaken. \textit{Test vs test} shows that tuning all models on the test set barely leads to a change in model rankings. For 4 of the 5 datasets, hyperparameter tuning allows for ample changes in model rankings, and comparing test-tuned models between themselves does not show evidence of overfitting.}
\label{tab:ranking-within-run}
\end{table}

\subsection{Exp 2: How much does test-tuning all models influence model selection?}

Now that we validated that our experimental setup allows for overfitting, we investigate if hyperparameter tuning all models on the test set leads to different model selection when compared to tuning hyperparameters on a validation set. 

In Table~\ref{tab:ranking-within-run} (left) we show the average number of model pair reorderings between our 30 runs when compared to validation tuning. We compare the deliberately overfitted (\textit{val vs test}) to tuning of all models on the test set (\textit{test vs test}). The \textit{test vs test} setting re-orders model less than the overfitted tuning.  
In Table~\ref{tab:ranking-within-run} (right) we show how many models are overtaking others when averaging over all runs. The overfitted setting (\textit{val vs test}) has 18 of 40 models change. Instead, when test tuning all models (\textit{test vs test}), only 2 models change rank.     

We further investigate  model selection by ranking models, as commonly done on a leaderboard. For each experiment, we fit test-tuned on validation-tuned test accuracy by ordinary least squares on the per-architecture means, and report Spearman and Kendall rank correlations on those same means. All confidence intervals are based on $100{,}000$ bootstrap resamples. For the linear fit we use a standard pairs bootstrap, resampling the eight architecture mean pairs with replacement. The rank correlations resample the $K$ runs within each architecture.

\begin{figure}[t]
    \centering
    \includegraphics[width=\linewidth]{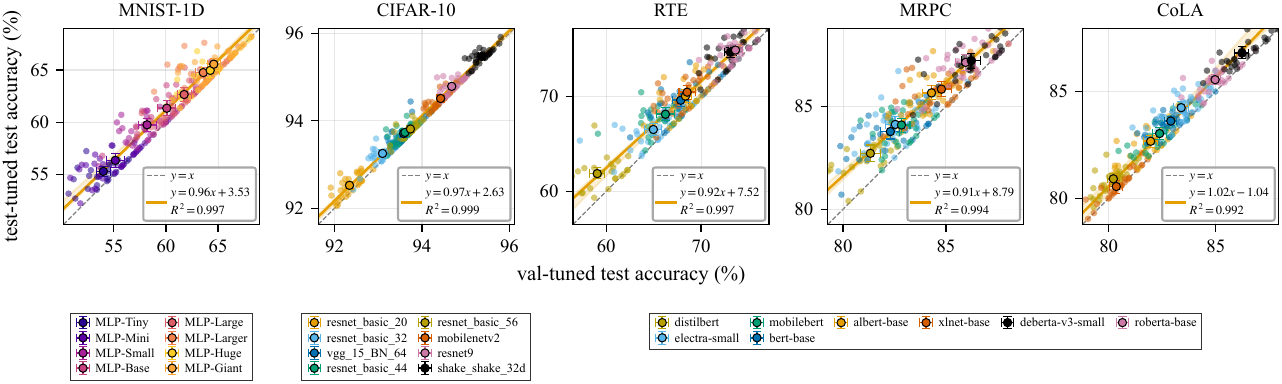}
\caption{Validation-tuned vs.\ test-tuned mean test accuracy. Markers are per-architecture means with $95\,\%$ bootstrap confidence intervals; dots are the individual paired runs. Color identifies the architecture, named in the legend below; the three GLUE tasks share one legend. The dashed grey line is the identity $y = x$ and the orange line is the ordinary least squares fit through the eight architecture means, with its $95\,\%$ pairs-bootstrap confidence interval shaded. On every benchmark the means lie close to a line nearly parallel to the identity, so test-tuning acts as a near-uniform upward shift that preserves architecture rankings.}
\label{fig:linear_fit}
\end{figure}

Figure~\ref{fig:linear_fit} shows the ordinary least squares fit between validation-tuned and test-tuned test accuracies, with per-architecture rankings in Table~\ref{tab:rq1_per_arch}. The slope is close to one ($0.91$--$1.02$) and the fit is tight ($R^2 \geq 0.99$). Test-tuning acts as a near-uniform upward shift and does not reorder the ranking. The architecture ranking is identical under both selection rules on MNIST-1D, CIFAR-10 and RTE (Spearman $\rho = 1.00$, Kendall $\tau = 1.00$). On MRPC and CoLA a single adjacent pair exchanges places ($\rho = 0.976$, $\tau = 0.929$), between architectures differing by $0.30\%$ against a run-to-run standard deviation of $1.34\%$ on MRPC, and by $0.14\%$ against $0.88\%$ on CoLA, so in both cases the two models are statistically indistinguishable under either rule. For all benchmarks and all models, adaptive overfitting does better in the \emph{absolute} accuracies, but not in the \emph{relative} comparison: a leaderboard built from val-tuned numbers selects the same winner as a leaderboard built from test-tuned numbers. We conclude that test tuning all models does not significantly change model rankings when compared to validation tuning all models.

\subsection{Exp 3: How inflated are test-tuning results compared to random variance?}

\begin{figure}
    \centering
    \includegraphics[width=1\linewidth]{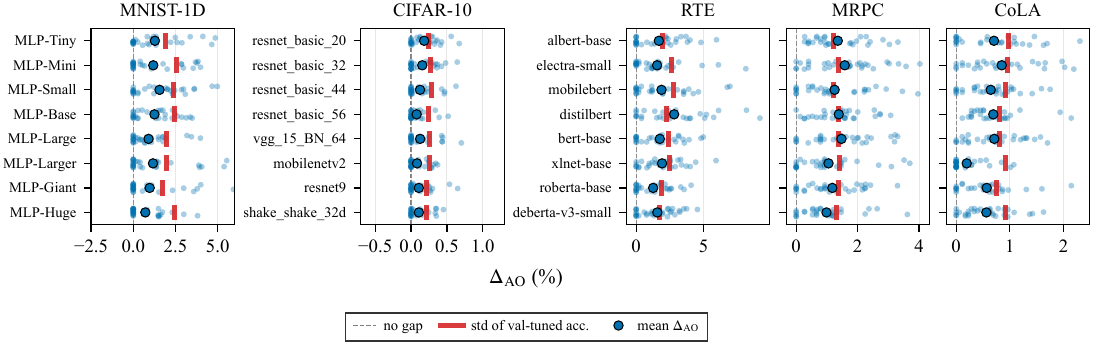}
        \caption{How does inflation by test-tuning compare to random variance: per-run adaptive overfitting gap, with architectures on the vertical axis. Each translucent dot is one of the $K = 30$ paired runs of an architecture; the filled blue circle is the per-architecture mean $\Delta_{\mathrm{AO}}$. The red vertical bar depicts one standard deviation of that architecture's val-tuned test accuracy, representing run-to-run noise. The MRPC and CoLA panels share the architecture labels of the RTE panel. On $35$ of the $40$ architecture--benchmark pairs the mean $\Delta_{\mathrm{AO}}$ sits below the standard deviation, showing that the gap is small relative to natural run-to-run variability, although test-tuning is always biased positively, it seems likely that overfitting by tuning random dataset splits instead of hyperparameters may allow higher inflation.  }
    \label{fig:per_run_gap_combined}
\end{figure}

Where our first experiment validated that our setting allows for overfitting, the second experiment showed that model rankings barely change when test tuning all models, but, their accuracy estimates are inflated. 
To evaluate the severity of this inflation, we compare the mean $\Delta_{\mathrm{OG}}$ over the $K{=}30$ paired runs against the natural run-to-run variability of the same architecture, measured as the standard deviation of the validation-tuned test accuracy across the same $K{=}30$ runs. If $\Delta_{\mathrm{OG}}$ is small relative to this experimental noise, the apparent overfitting is no larger than what re-running the pipeline under different data splits and initializations would already produce.

Results in Figure~\ref{fig:per_run_gap_combined} show that on $35$ of the $40$ architecture--benchmark pairs, the mean gap is smaller than the run-to-run standard deviation of the validation-tuned models. The margin is widest on the image benchmarks, where the gap is roughly half the standard deviation and never exceeds $0.74$ of it, and on CoLA, where it never exceeds $0.89$. The five exceptions are four of the eight MRPC architectures, at up to $1.13$, and \texttt{distilbert} on RTE at $1.25$, the largest ratio observed anywhere. We conclude that overfitting severity, in our setup, is comparable to the natural experimental variability in the experimental design. We also checked whether the gap concentrates in a single hyperparameter. Averaged over the architectures of a benchmark, the learning rate is the largest fANOVA main effect everywhere but reaches only $24$--$32\%$ of the total importance, and the initialisation seed accounts for $16$--$28\%$, so test-tuning is not simply selecting a lucky initialisation (Appendix~\ref{app:fanova}).

\begin{table}[t]
  \centering
  \scriptsize
  \begin{tabular}{r l r r r r r r}
    \toprule
    Val. & Model & Validation-tuned & Test-tuned & $\Delta_{\mathrm{OG}}$ & Val. Std. & $\Delta$Rank &  $\Delta$Rank \\
    Rank &       & (\%)             & (\%)       &                        & (\%)      &  Val vs Test &  Test-only \\
    \midrule
    \multicolumn{8}{c}{\textbf{MNIST-1D}} \\
    \midrule
     1 & \texttt{MLP-Giant}  & 64.59 & 65.56 & $+0.97$ & $1.76$ & 0 & $\phantom{+}0$ \\
     2 & \texttt{MLP-Huge}   & 64.25 & 64.96 & $+0.72$ & $2.46$ & +1  & $\phantom{+}0$ \\
     3 & \texttt{MLP-Larger} & 63.58 & 64.75 & $+1.17$ & $1.97$ & +2 & $\phantom{+}0$ \\
     4 & \texttt{MLP-Large}  & 61.74 & 62.65 & $+0.91$ & $1.96$ & 0 & $\phantom{+}0$ \\
     5 & \texttt{MLP-Base}   & 60.08 & 61.33 & $+1.26$ & $2.44$ & 0 & $\phantom{+}0$ \\
     6 & \texttt{MLP-Small}  & 58.19 & 59.74 & $+1.56$ & $2.41$ & 0 & $\phantom{+}0$ \\
     7 & \texttt{MLP-Mini}   & 55.15 & 56.33 & $+1.19$ & $2.54$ & 0 & $\phantom{+}0$ \\
     8 & \texttt{MLP-Tiny}   & 53.99 & 55.28 & $+1.28$ & $1.92$ & +1 & $\phantom{+}0$ \\
    \midrule
    \multicolumn{8}{c}{\textbf{CIFAR-10}} \\
    \midrule
     1 & \texttt{shake\_shake\_32d}  & 95.36 & 95.47 & $+0.11$ & $0.21$ & 0 & $\phantom{+}0$ \\
     2 & \texttt{resnet9}            & 94.68 & 94.79 & $+0.11$ & $0.22$ & 0 & $\phantom{+}0$ \\
     3 & \texttt{mobilenetv2}        & 94.43 & 94.51 & $+0.08$ & $0.26$ & 0 & $\phantom{+}0$ \\
     4 & \texttt{resnet\_basic\_56}  & 93.74 & 93.81 & $+0.08$ & $0.24$ & 0 & $\phantom{+}0$ \\
     5 & \texttt{resnet\_basic\_44}  & 93.61 & 93.73 & $+0.12$ & $0.29$ & 0 & $\phantom{+}0$ \\
     6 & \texttt{vgg\_15\_BN\_64}    & 93.58 & 93.71 & $+0.13$ & $0.26$ & +1 & $\phantom{+}0$ \\
     7 & \texttt{resnet\_basic\_32}  & 93.10 & 93.25 & $+0.16$ & $0.28$ & 0 & $\phantom{+}0$ \\
     8 & \texttt{resnet\_basic\_20}  & 92.34 & 92.52 & $+0.18$ & $0.25$ & 0 & $\phantom{+}0$ \\
    \midrule
    \multicolumn{8}{c}{\textbf{RTE}} \\
    \midrule
     1 & \texttt{roberta-base}     & 73.62 & 74.88 & $+1.26$ & $1.88$ & 0 & $\phantom{+}0$ \\
     2 & \texttt{deberta-v3-small} & 73.08 & 74.65 & $+1.57$ & $1.75$ & +1 & $\phantom{+}0$ \\
     3 & \texttt{xlnet-base}       & 68.53 & 70.45 & $+1.92$ & $2.50$ & 0 & $\phantom{+}0$ \\
     4 & \texttt{albert-base}      & 68.35 & 70.03 & $+1.68$ & $1.92$ & +1 & $\phantom{+}0$ \\
     5 & \texttt{bert-base}        & 67.85 & 69.60 & $+1.75$ & $2.43$ & +2 & $\phantom{+}0$ \\
     6 & \texttt{mobilebert}       & 66.24 & 68.12 & $+1.88$ & $2.73$ & +1 & $\phantom{+}0$ \\
     7 & \texttt{electra-small}    & 64.97 & 66.52 & $+1.55$ & $2.61$ & +1 & $\phantom{+}0$ \\
     8 & \texttt{distilbert}       & 59.06 & 61.87 & $+2.81$ & $2.25$ & 0 & $\phantom{+}0$ \\
    \midrule
    \multicolumn{8}{c}{\textbf{MRPC}} \\
    \midrule
     1 & \texttt{deberta-v3-small} & 86.23 & 87.22 & $+0.99$ & $1.33$ & 0 & $\phantom{+}0$ \\
     2 & \texttt{roberta-base}     & 85.96 & 87.13 & $+1.18$ & $1.36$ & +1 & $\phantom{+}0$ \\
     3 & \texttt{xlnet-base}       & 84.78 & 85.83 & $+1.05$ & $1.41$ & 0 & $\phantom{+}0$ \\
     4 & \texttt{albert-base}      & 84.30 & 85.65 & $+1.35$ & $1.21$ & +1 & $\phantom{+}0$ \\
     5 & \texttt{mobilebert}       & 82.84 & 84.09 & $+1.25$ & $1.22$ & 0 & $-1$ \\
     6 & \texttt{electra-small}    & 82.54 & 84.12 & $+1.58$ & $1.39$ & +1 & $+1$ \\
     7 & \texttt{bert-base}        & 82.30 & 83.77 & $+1.47$ & $1.37$ & +2 & $\phantom{+}0$ \\
     8 & \texttt{distilbert}       & 81.33 & 82.71 & $+1.39$ & $1.39$ & +2 & $\phantom{+}0$ \\
    \midrule
    \multicolumn{8}{c}{\textbf{CoLA}} \\
    \midrule
     1 & \texttt{deberta-v3-small} & 86.22 & 86.78 & $+0.56$ & $0.93$ & 0 & $\phantom{+}0$ \\
     2 & \texttt{roberta-base}     & 84.97 & 85.54 & $+0.57$ & $0.76$ & 0 & $\phantom{+}0$ \\
     3 & \texttt{electra-small}    & 83.38 & 84.23 & $+0.85$ & $0.96$ & 0 & $\phantom{+}0$ \\
     4 & \texttt{bert-base}        & 82.90 & 83.61 & $+0.71$ & $0.81$ & +1 & $\phantom{+}0$ \\
     5 & \texttt{mobilebert}       & 82.38 & 83.02 & $+0.64$ & $0.91$ & +1 & $\phantom{+}0$ \\
     6 & \texttt{albert-base}      & 81.96 & 82.66 & $+0.70$ & $0.97$ & +1 & $\phantom{+}0$ \\
     7 & \texttt{xlnet-base}       & 80.36 & 80.55 & $+0.20$ & $0.91$ & 0 & $-1$ \\
     8 & \texttt{distilbert}       & 80.22 & 80.91 & $+0.69$ & $0.81$ & +1 & $+1$ \\
    \bottomrule
  \end{tabular}
  \caption{Model rankings ordered by validation-tuned accuracy, with the overfitting difference and the run-to-run standard deviation of validation-tuned accuracy. Several models within a benchmark are close together and have near-similar accuracy. Our experimental setup allows to evalute overfitting by hyperparameter tuning because when validation tuned models are compared to test-tuned models, there are ample ranking changes (Val vs Test). For Test-only, on MNIST-1D, CIFAR-10 and RTE no model changes rank when tuned on the test set rather than the validation set. On MRPC and CoLA exactly one adjacent pair exchanges places, in both cases between models separated by far less than the run-to-run standard deviation, so they are statistically indistinguishable under either rule. The top-ranked model is unchanged on every benchmark.}
  \label{tab:rq1_per_arch}
\end{table}

\section{Discussion and Limitations}

The experiments tell a consistent story across four of our five benchmarks, two modalities and three model families. It is easy to overfit by tuning on the test set, but the severity stays within the range of natural experimental variability such as random dataset splits and training randomness. The mean gap is roughly half the run-to-run standard deviation on the image benchmarks and, on average, between $0.7$ and $1.0$ times it on the GLUE tasks. Test-tuning yields a consistent absolute improvement without reordering models: rankings are identical under both selection rules on MNIST-1D and RTE, and on MRPC and CoLA the only change is a single adjacent inversion between architectures separated by far less than run-to-run noise. The top-ranked model is unchanged everywhere.

The results agree with the empirical literature. \citet{recht2018cifar, recht2019imagenet_generalize} and~\citet{roelofs2019meta_overfitting} both find that adaptive overfitting on popular benchmarks is small and does not change model rankings. Our work extends this by measuring explicit test set overfitting for all models in a model selection setting. In practice, this means that reusing the test set for hyperparameter tuning under random search at this budget produces a measurable but small accuracy inflation, on image classification and natural language understanding alike.

\subsection{Limitations}

\paragraph{Random search is a weak adaptive querier.}
The methodology uses random search, which evaluates each hyperparameter configuration independently. We use this experimental setup because this is the setting for which we have evidence that test-tuning occurs in the literature. More advanced methods such as Bayesian optimization, hyperband~\citet{li2018hyperband}, or manual iterative tuning reuse results from earlier evaluations, could produce larger overfitting gaps. We have not seen these settings used in practice for test-tuning, but would be interesting future research.

\paragraph{Limited domain coverage.}
The larger GLUE tasks, other modalities, and regression are not tested. \citet{liao2021learning_eval_failures} suggest that  overfitting may be more severe for regression. We are not aware of existing regression work that tunes on the test-set, yet, it would be interesting future work.

\paragraph{Wide confidence intervals.}
$K{=}30$ paired runs is the minimum recommended by \citet{bouthillier2021variance} via the Noether bound~\citep{noether1987sample}, so the bootstrap intervals around $\hat{P}(\Delta_{\mathrm{AO}} > 0)$ are wide, around $0.10$--$0.17$. The weakest architectures have lower bounds near $0.60$, so their individual results are not robust, but the overall pattern of small positive gaps does not depend on them.

\paragraph{Only visual and textual models evaluated.} We only evaluate common deep learning architectures, limited to the image and text domain. We did not explore other machine learning algorithms.

\paragraph{Misinterpretation risks.}  We do not aim to normalize test-tuning. Test-tuning leads to inflated accuracy estimates, and our results confirm that comparing validation-tuned models to test-tuned models is unfair. Our results do bring nuance. Not all test-tuning is flawed, and our results suggest that if new research needs to compare to existing test-tuned research, that this comparison can be meaningful.

\section{Conclusion}

We presented a paired-run methodology for directly measuring  overfitting from hyperparameter tuning, applied to a controlled MNIST-1D family of $8$ MLPs, a realistic suite of $8$ CIFAR-10 architectures, and three small GLUE tasks each fine-tuning $8$ pretrained transformers, for $36{,}000$ trained models in total. We investigate how easy and severe test-tuning is and show  overfitting on all $40$ architecture--benchmark combinations. The effect was small in absolute terms, and within the range of the natural experimental variation. We also investigate whether test-tuning changes which model would be selected, where it behaved as a near-uniform upward shift: the linear fit through the per-architecture means had slope close to one ($0.91$--$1.02$) and $R^2 \geq 0.99$ on every benchmark, and rank correlations were exactly $1$ on MNIST-1D, CIFAR-10 and RTE. On MRPC and CoLA a single adjacent pair exchanged places, in both cases between architectures separated by far less than run-to-run noise, and the top-ranked model was unchanged everywhere. Taken together, the experiments suggest that for honest random-search hyperparameter tuning, the practical risk of test-set tuning is  smaller than worst-case theory predicts, and comparing accuracies under test-tuning does not change which model a leaderboard would select. We do not aim to normalize test-tuning, but if it cannot be avoided, then these empirical results allow researchers to discuss, mention, and defend tuning hyperparameters on the test set, leading to a more open scientific discourse.


\subsection*{AI use statement}

In this work, we used generative AI tools for coding support, such as refactoring and debugging. We have not used generative AI tools for generating research ideas, designing experiments, producing results, or creating data, and other required disclosure tasks are not applicable to this work. Additionally, we used generative AI tools to improve the clarity of the writing and to correct grammatical errors. We have reviewed all AI-assisted work. AI-assisted code was manually verified and tested for correctness, and all AI-suggested edits to the writing were checked against the original intent and meaning before acceptance. We take responsibility for the final content of this work, including text, claims, or artifacts produced with the aid of generative AI.

\subsection*{Ethics statement}

This work studies a methodological issue in machine learning evaluation, namely how strongly tuning hyperparameters on the test set inflates reported accuracy. It does not introduce new models or datasets, and we do not anticipate direct negative societal consequences from its release. If anything, a clearer empirical picture of when test-set tuning matters and when it does not should help researchers and reviewers assess published results more accurately, reducing the risk that benchmark numbers are taken at face value without questioning the selection protocol behind them. All experiments are run on standard public datasets (MNIST-1D, CIFAR-10 and three GLUE tasks) on shared academic cluster hardware, and the released code and analysis scripts are intended to make the findings independently verifiable.

\subsection*{Reproducibility statement}
\label{reproducibility}

All code, experiment configurations, and analysis scripts required to reproduce our results are available at \url{https://anonymous.4open.science/r/hyperparameter-tuning-on-test-set/README.md}. All randomness derives from a single master seed: NumPy's \texttt{SeedSequence} splits it into independent sub-seeds, one per architecture, per run, and per source of randomness within a run (bootstrap split, hyperparameter sampling -- including the per-trial initialisation seed -- and other training stochasticity). No two sub-seeds coincide, so the full experiment is reproducible end-to-end from that single seed. The paired-run protocol, the two selection rules, and the definition of $\Delta_{\mathrm{AO}}$ are given in Section~\ref{sec:methodology}, and the experimental settings, budgets, and compute environment in Section~\ref{sec:experiments}. Dataset construction and the out-of-bootstrap splitting scheme are detailed in Appendix~\ref{app:data-splitting}, the architectures and their reference implementations in Appendix~\ref{app:architectures}, the hyperparameter search spaces in Appendix~\ref{app:hpo-spaces}, and the training setup in Appendix~\ref{app:implementation-details}. Appendix~\ref{app:ranking-details} reports the exact bootstrap settings and full confidence intervals behind the ranking analysis, and Appendix~\ref{app:hp-validity} the empirical winner distributions used to verify that the search ranges are well-shaped. The supporting analyses are likewise fully specified: the memorisation upper bound in Appendix~\ref{app:upper-bound}, the larger-budget MLP-Giant$_{T=100}$ rerun in Appendix~\ref{app:robustness}, and the CIFAR-10.1 transfer evaluation in Appendix~\ref{app:cifar101}.

\bibliography{article}
\bibliographystyle{iclr2027_conference}

\appendix
\clearpage
\appendix

\FloatBarrier
\section{Runs with identical configurations}
\label{app:glue-extra}

\begin{table}[t]
  \centering
  \small
  \begin{tabular}{lccc}
    \toprule
    Benchmark & $\Delta_{\mathrm{AO}} = 0$ & Percentage & Identical configuration \\
    \midrule
    MNIST-1D & $99 / 240$  & $41.2\%$ & $98.0\%$ \\
    CIFAR-10 & $116 / 240$ & $48.3\%$ & $98.3\%$ \\
    MRPC     & $31 / 240$  & $12.9\%$ & $90.3\%$ \\
    \bottomrule
  \end{tabular}
  \caption{Runs with $\Delta_{\mathrm{AO}} = 0$ pooled over the eight architectures of each benchmark ($K = 30$ runs each), as a count and as a percentage of all runs. The last column is the share of runs with $\Delta_{\mathrm{AO}} = 0$ in which validation-tuning and test-tuning selected the identical configuration. The remainder are distinct configurations that happen to score identically on the test split. RTE and CoLA are in Appendix~\ref{app:glue-extra}.  }
  \label{tab:zero_gap_combined}
\end{table}

A substantial fraction of runs have $\Delta_{\mathrm{AO}} = 0$: validation- and test-tuning select the same configuration, so test-tuning gains nothing. Table~\ref{tab:zero_gap_combined} shows this fraction is high on the image benchmarks ($41.2\%$ on MNIST-1D, $48.3\%$ on CIFAR-10) and lower on MRPC ($12.9\%$), which is also where the gap is largest relative to run-to-run noise. On every benchmark this effect is overwhelmingly caused by both selection strategies picking the same hyperparameter configurations instead of accidental ties between distinct configurations. When one configuration wins on both splits, test-tuning has nothing to gain.

\section{Implementation Details}
\label{app:implementation-details}

\paragraph{Batch size.} The two image experiments use a training batch size of $128$ and an evaluation batch size of $512$. The GLUE experiment uses a training batch size of $32$, following the published fine-tuning recipes of the model families we evaluate, and an evaluation batch size of $128$.

\paragraph{Data augmentation (CIFAR-10).} During training, each mini-batch is augmented with random cropping ($32 \times 32$ with $4$-pixel padding) and random horizontal flipping, followed by per-channel normalisation to zero mean and unit variance. At evaluation time, only the normalisation is applied.

\paragraph{Mixed-precision training (CIFAR-10).} All CIFAR-10 models are trained with automatic mixed precision (\texttt{torch.amp}) using a gradient scaler to maintain numerical stability.

\paragraph{Training budget (CIFAR-10).} The number of training steps is itself a tuned hyperparameter, sampled log-uniformly from $[75{,}000,\; 141{,}000]$. At batch size $128$, the bootstrapped $60{,}000$-example training split yields ${\approx}\,470$ mini-batches per epoch, so this range corresponds to roughly $160$--$300$ epochs. This sits within the typical training-length range reported for most of the architectures we use, providing enough budget for each configuration to converge while keeping experiments computationally manageable. Since our goal is to measure adaptive overfitting rather than to reproduce state-of-the-art accuracies, what matters is that the hyperparameter choices yield meaningful accuracy differences between configurations.

\paragraph{Fine-tuning setup (GLUE).} Each GLUE trial loads a fresh pretrained checkpoint with a randomly initialised classification head and fine-tunes all parameters; no layers are frozen. Inputs are tokenised with each model's own tokenizer, truncated and padded to a maximum sequence length of $128$. Training uses AdamW with a linear warmup-then-decay schedule and gradient clipping at a maximum norm of $1.0$. The number of epochs, the warmup ratio, the learning rate and the weight decay are all tuned (Table~\ref{tab:hpo-spaces-glue}); the batch size is held fixed at $32$. The initialisation seed is sampled per trial and applied \emph{before} the model is constructed, so the classification head's initialisation and the data ordering are both part of the trial's seed --- fine-tuning on small GLUE tasks is known to be highly sensitive to this choice~\cite{dodge2020fine}.

\paragraph{Numerical precision (GLUE).} All GLUE models are trained with automatic mixed precision and a gradient scaler, with one exception: MobileBERT's activations overflow in fp16, so it is trained in full fp32. Checkpoints are loaded with an explicit fp32 master dtype, because some checkpoints (e.g.\ DeBERTa-v3-small) declare a half-precision \texttt{torch\_dtype} and would otherwise load in fp16, which is incompatible with gradient scaling. With fp32 master weights, autocast still casts activations to fp16 in the forward pass while gradients remain fp32.

\FloatBarrier
\section{Hyperparameter Search Spaces}
\label{app:hpo-spaces}

\begin{table}[!htbp]
\caption{Hyperparameter search space for MNIST-1D.}
\label{tab:hpo-spaces-mnist1d}
\centering
\begin{small}
\begin{tabular}{llc}
\toprule
\textbf{Hyperparameter} & \textbf{Distribution} & \textbf{Range} \\
\midrule
Learning rate            & Log-Uniform           & $[10^{-4}, 10^{-1}]$ \\
Learning-rate decay rate & Log-Uniform$^\dagger$ & $[10^{-5}, 10^{-2}]$ \\
Weight decay             & Log-Uniform           & $[10^{-6}, 10^{-2}]$ \\
Initialization seed      & Uniform integer       & $[0,\; 2^{31}-1]$ \\
\bottomrule
\multicolumn{3}{l}{\footnotesize $^\dagger$ Set to zero with probability $0.3$ to explicitly sample the no-decay setting.}
\end{tabular}
\end{small}
\end{table}

\begin{table}[!htbp]
\caption{Hyperparameter search space for CIFAR-10.}
\label{tab:hpo-spaces-cifar10}
\centering
\begin{small}
\begin{tabular}{llc}
\toprule
\textbf{Hyperparameter} & \textbf{Distribution} & \textbf{Range} \\
\midrule
Learning rate       & Log-Uniform     & $[5 \times 10^{-3},\; 0.5]$ \\
Weight decay        & Log-Uniform     & $[10^{-5},\; 10^{-2}]$      \\
SGD momentum        & Uniform         & $[0.8,\; 0.99]$             \\
Training steps      & Log-Uniform     & $[75{,}000,\; 141{,}000]$   \\
Initialization seed & Uniform integer & $[0,\; 2^{31}-1]$           \\
\bottomrule
\end{tabular}
\end{small}
\end{table}

\begin{table}[!htbp]
\caption{Hyperparameter search space for the GLUE tasks, shared across all three tasks and all eight architectures. The ranges bracket the published fine-tuning grids of the model families in the registry: learning rates of $2$--$5 \times 10^{-5}$ (BERT, XLNet), $1$--$3 \times 10^{-5}$ (RoBERTa), $1$--$5 \times 10^{-5}$ (ALBERT) and $10^{-4}$ (ELECTRA-base); weight decays of $0$ (HuggingFace \texttt{run\_glue} default), $0.01$ (BERT, DeBERTa) and $0.1$ (RoBERTa); warmup ratios of $6\%$ (RoBERTa) and $10\%$ (BERT); and epoch counts of $3$ (BERT) through $10$ (RoBERTa).}
\label{tab:hpo-spaces-glue}
\centering
\begin{small}
\begin{tabular}{llc}
\toprule
\textbf{Hyperparameter} & \textbf{Distribution} & \textbf{Range} \\
\midrule
Learning rate       & Log-Uniform     & $[5 \times 10^{-6},\; 10^{-4}]$ \\
Weight decay        & Log-Uniform     & $[10^{-5},\; 10^{-1}]$          \\
Warmup ratio        & Uniform         & $[0.0,\; 0.2]$                  \\
Number of epochs    & Uniform integer & $[2,\; 10]$                     \\
Initialization seed & Uniform integer & $[0,\; 2^{31}-1]$               \\
\bottomrule
\end{tabular}
\end{small}
\end{table}

\pagebreak

\FloatBarrier
\section{Architectures}
\label{app:architectures}

\needspace{12\baselineskip}
\subsection{MNIST-1D}

\begin{table}[!htbp]
\caption{MLP family evaluated on MNIST-1D. All networks use ReLU activations and residual connections between hidden layers.}
\label{tab:mnist1d-architectures}
\centering
\begin{small}
\begin{tabular}{lcr}
\toprule
\textbf{Architecture} & \textbf{Hidden sizes} & \textbf{Parameters} \\
\midrule
MLP-Tiny    & $[25]$                 & $1{,}285$   \\
MLP-Mini    & $[50]$                 & $2{,}560$   \\
MLP-Small   & $[100]$                & $5{,}110$   \\
MLP-Base    & $[100, 100]$           & $15{,}210$  \\
MLP-Large   & $[100, 100, 100]$      & $25{,}310$  \\
MLP-Larger  & $[200, 200, 200]$      & $90{,}610$  \\
MLP-Giant   & $[200, 200, 200, 200]$ & $130{,}810$ \\
MLP-Huge    & $[300, 300, 300]$      & $195{,}910$ \\
\bottomrule
\end{tabular}
\end{small}
\end{table}

\FloatBarrier
\needspace{12\baselineskip}
\subsection{CIFAR-10}

\begin{table}[!htbp]
\caption{Architectures evaluated on CIFAR-10. Each model is ported from its reference implementation without architectural modifications.}
\label{tab:cifar-architectures}
\centering
\small
\begin{tabular}{l l l l}
\toprule
\# & Model & Paper & Repository \\
\midrule
1 & ResNet-20         & \cite{he2016deep}             & \url{github.com/hysts/pytorch_image_classification} \\
2 & ResNet-32         & \cite{he2016deep}             & \url{github.com/hysts/pytorch_image_classification} \\
3 & ResNet-44         & \cite{he2016deep}             & \url{github.com/hysts/pytorch_image_classification} \\
4 & ResNet-56         & \cite{he2016deep}             & \url{github.com/hysts/pytorch_image_classification} \\
5 & Shake-Shake (32d) & \cite{gastaldi2017shake}      & \url{github.com/hysts/pytorch_image_classification} \\
6 & VGG-15 BN         & \cite{simonyan2014very}       & \url{github.com/hysts/pytorch_image_classification} \\
7 & MobileNetV2       & \cite{sandler2018mobilenetv2} & \url{github.com/hysts/pytorch_image_classification} \\
8 & ResNet-9          & \cite{page2019how}            & \url{github.com/KellerJordan/cifar10-airbench} \\
\bottomrule
\end{tabular}
\end{table}

\FloatBarrier
\needspace{12\baselineskip}
\subsection{GLUE}

\begin{table}[!htbp]
\caption{Pretrained transformers fine-tuned on each GLUE task. All checkpoints are public Hugging Face models, used without architectural modification; each trial loads the checkpoint afresh and attaches a randomly initialised classification head. The same eight models are evaluated on all three tasks.}
\label{tab:glue-architectures}
\centering
\small
\begin{tabular}{l l l r l}
\toprule
\# & Model & Paper & Parameters & Hugging Face checkpoint \\
\midrule
1 & ALBERT-base-v2    & \cite{lan2020albert}     & $11.8$M  & \texttt{albert/albert-base-v2} \\
2 & ELECTRA-small     & \cite{clark2020electra}  & $14.0$M  & \texttt{google/electra-small-discriminator} \\
3 & MobileBERT        & \cite{sun2020mobilebert} & $25.3$M  & \texttt{google/mobilebert-uncased} \\
4 & DistilBERT-base   & \cite{sanh2019distilbert}& $66.4$M  & \texttt{distilbert/distilbert-base-uncased} \\
5 & BERT-base         & \cite{devlin2019bert}    & $109.5$M & \texttt{google-bert/bert-base-uncased} \\
6 & XLNet-base        & \cite{yang2019xlnet}     & $110.0$M & \texttt{xlnet/xlnet-base-cased} \\
7 & RoBERTa-base      & \cite{liu2019roberta}    & $124.6$M & \texttt{FacebookAI/roberta-base} \\
8 & DeBERTa-v3-small  & \cite{he2021debertav3}   & $141.9$M & \texttt{microsoft/deberta-v3-small} \\
\bottomrule
\end{tabular}
\end{table}

\FloatBarrier
\section{Ranking Preservation Details}
\label{app:ranking-details}
\FloatBarrier

For each benchmark we fit test-tuned on validation-tuned test accuracy by ordinary least squares on the per-architecture means, and report Spearman and Kendall rank correlations on those same means. All confidence intervals are based on bootstrap resampling. For the linear fit we use a standard pairs bootstrap with $100{,}000$ resamples, drawing the eight architecture mean pairs with replacement and refitting the line at each iteration. The rank correlations use a cluster bootstrap with $20{,}000$ resamples, drawing the $K$ runs within each architecture with replacement and recomputing the architecture means at each iteration.

\subsection{MNIST-1D}
Linear fit:
\begin{itemize}
  \item Slope $= 0.960$, \; $95\,\%$ bootstrap CI: $[0.893,\; 0.988]$
  \item Intercept $= 3.534$, \; $95\,\%$ bootstrap CI: $[1.885,\; 7.708]$
  \item $R^2 = 0.997$, \; $95\,\%$ bootstrap CI: $[0.992,\; 1.000]$
  \item $p$-value $= 5.73 \times 10^{-9}$
\end{itemize}
Rank correlations:
\begin{itemize}
  \item Spearman $\rho = 1.000$ \; ($p \approx 0$), \; $95\,\%$ bootstrap CI: $[0.929,\; 1.000]$
  \item Kendall $\tau = 1.000$ \; ($p = 4.96 \times 10^{-5}$), \; $95\,\%$ bootstrap CI: $[0.857,\; 1.000]$
\end{itemize}

\subsection{CIFAR-10}
Linear fit:
\begin{itemize}
  \item Slope $= 0.973$, \; $95\,\%$ bootstrap CI: $[0.942,\; 1.001]$
  \item Intercept $= 2.631$, \; $95\,\%$ bootstrap CI: $[0.043,\; 5.544]$
  \item $R^2 = 0.999$, \; $95\,\%$ bootstrap CI: $[0.998,\; 1.000]$
  \item $p$-value $= 1.08 \times 10^{-10}$
\end{itemize}
Rank correlations:
\begin{itemize}
  \item Spearman $\rho = 1.000$ \; ($p \approx 0$), \; $95\,\%$ bootstrap CI: $[0.976,\; 1.000]$
  \item Kendall $\tau = 1.000$ \; ($p = 4.96 \times 10^{-5}$), \; $95\,\%$ bootstrap CI: $[0.929,\; 1.000]$
\end{itemize}

\subsection{RTE}
Linear fit:
\begin{itemize}
  \item Slope $= 0.916$, \; $95\,\%$ bootstrap CI: $[0.885,\; 1.007]$
  \item Intercept $= 7.523$, \; $95\,\%$ bootstrap CI: $[1.202,\; 9.531]$
  \item $R^2 = 0.997$, \; $95\,\%$ bootstrap CI: $[0.992,\; 1.000]$
  \item $p$-value $= 1.06 \times 10^{-8}$
\end{itemize}
Rank correlations:
\begin{itemize}
  \item Spearman $\rho = 1.000$ \; ($p \approx 0$), \; $95\,\%$ bootstrap CI: $[0.905,\; 1.000]$
  \item Kendall $\tau = 1.000$ \; ($p = 4.96 \times 10^{-5}$), \; $95\,\%$ bootstrap CI: $[0.786,\; 1.000]$
\end{itemize}

\subsection{MRPC}
Linear fit:
\begin{itemize}
  \item Slope $= 0.910$, \; $95\,\%$ bootstrap CI: $[0.850,\; 0.960]$
  \item Intercept $= 8.787$, \; $95\,\%$ bootstrap CI: $[4.610,\; 13.889]$
  \item $R^2 = 0.994$, \; $95\,\%$ bootstrap CI: $[0.982,\; 0.999]$
  \item $p$-value $= 6.58 \times 10^{-8}$
\end{itemize}
Rank correlations:
\begin{itemize}
  \item Spearman $\rho = 0.976$ \; ($p = 3.31 \times 10^{-5}$), \; $95\,\%$ bootstrap CI: $[0.905,\; 1.000]$
  \item Kendall $\tau = 0.929$ \; ($p = 3.97 \times 10^{-4}$), \; $95\,\%$ bootstrap CI: $[0.786,\; 1.000]$
\end{itemize}
The single rank change is an exchange between \texttt{mobilebert} and \texttt{electra-small}, whose validation-tuned means differ by $0.30\%$ against a run-to-run standard deviation of $1.34\%$.

\subsection{CoLA}
Linear fit:
\begin{itemize}
  \item Slope $= 1.020$, \; $95\,\%$ bootstrap CI: $[0.956,\; 1.157]$
  \item Intercept $= -1.040$, \; $95\,\%$ bootstrap CI: $[-12.260,\; 4.374]$
  \item $R^2 = 0.992$, \; $95\,\%$ bootstrap CI: $[0.985,\; 1.000]$
  \item $p$-value $= 1.47 \times 10^{-7}$
\end{itemize}
Rank correlations:
\begin{itemize}
  \item Spearman $\rho = 0.976$ \; ($p = 3.31 \times 10^{-5}$), \; $95\,\%$ bootstrap CI: $[0.952,\; 1.000]$
  \item Kendall $\tau = 0.929$ \; ($p = 3.97 \times 10^{-4}$), \; $95\,\%$ bootstrap CI: $[0.857,\; 1.000]$
\end{itemize}
The single rank change is an exchange between \texttt{xlnet-base} and \texttt{distilbert}, whose validation-tuned means differ by $0.14\%$ against a run-to-run standard deviation of $0.88\%$.

\section{Upper-Bound Reference}
\label{app:upper-bound}

\begin{figure}[!htbp]
  \centering
  \includegraphics[width=1\linewidth]{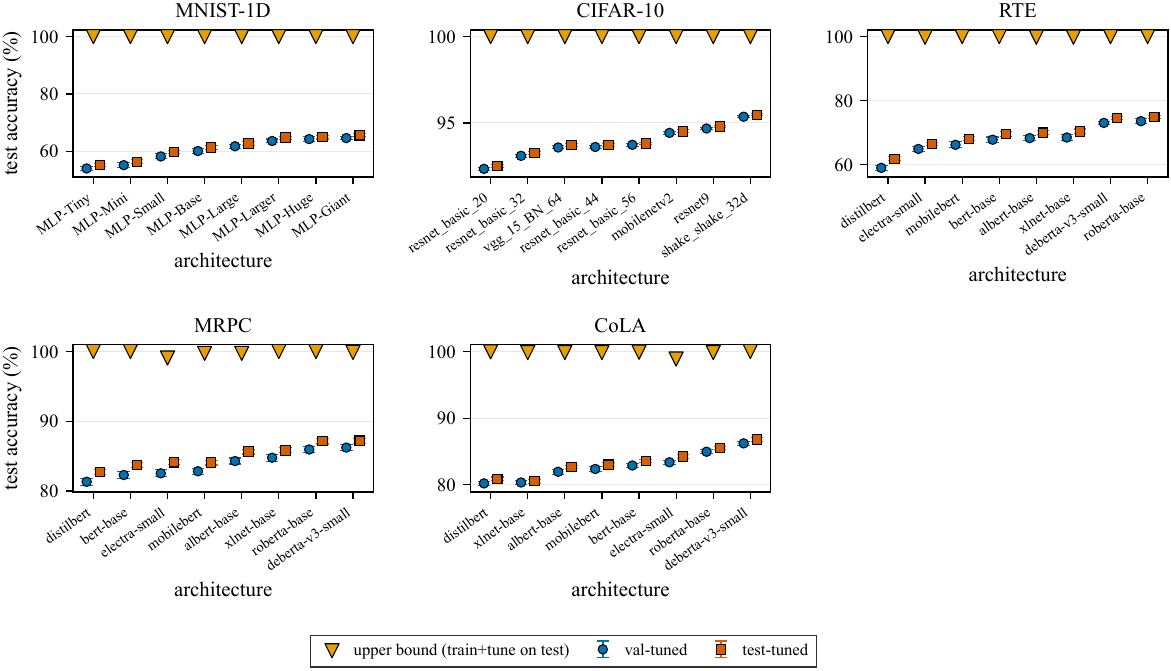}
  \caption{Upper-bound reference per architecture across the five benchmarks, ordered by mean validation-tuned test accuracy. Blue circles are the validation-tuned mean test accuracy and orange squares the test-tuned mean test accuracy (both with $95\,\%$ bootstrap CIs over the $K = 30$ runs). Orange triangles are the upper bound: the best test accuracy among $T = 30$ models trained directly on the same bootstrap test fold. The vertical distance between the blue marker and the orange triangle is the headroom available to any procedure that exploits the test set; the much smaller distance between the blue circle and orange square is the share that random search over $T = 30$ trials actually captures.}
  \label{fig:upper_bound_combined}
\end{figure}

\begin{table}[!htbp]
  \centering
  \small
  \begin{tabular}{lccc}
    \toprule
    Benchmark & Upper bound (\%) & Headroom (\%) & Captured (\%) \\
    \midrule
    MNIST-1D & $100.00$          & $35.4$--$46.0$ & $2.0$--$3.7$ \\
    CIFAR-10 & $100.00$          & $4.6$--$7.7$   & $1.2$--$2.4$ \\
    RTE      & $99.80$--$100.00$ & $26.4$--$40.9$ & $4.4$--$6.9$ \\
    MRPC     & $99.06$--$100.00$ & $13.6$--$18.7$ & $6.9$--$9.6$ \\
    CoLA     & $98.87$--$100.00$ & $13.8$--$19.7$ & $1.0$--$5.5$ \\
    \bottomrule
  \end{tabular}
  \caption{Ranges over the eight architectures of each benchmark. \emph{Upper bound} is the best test accuracy among $T = 30$ models trained and tuned directly on the bootstrap test fold. \emph{Headroom} is the distance from the validation-tuned mean to that bound, and \emph{captured} is the share of it that test-tuning actually realises, $(A(\lambda^*_{\text{test}}) - A(\lambda^*_{\text{val}})) / \text{headroom}$.}
  \label{tab:upper_bound_ranges}
\end{table}

Every architecture very nearly memorises its own test fold: exactly $100\%$ on both image benchmarks, and between $98.9\%$ and $100\%$ on the GLUE tasks, whose bootstrap test folds are smaller (${\sim}509$ examples on RTE to ${\sim}1{,}763$ on CoLA). Even MLP-Tiny has enough capacity to memorise the ${\sim}736$-sample MNIST-1D fold. Headroom above the validation-tuned mean varies widely across benchmarks, from $4.6$--$7.7\%$ on CIFAR-10, where validation-tuned accuracy is already in the low-$90\%$s, up to $35$--$46\%$ on MNIST-1D. Yet in every case random search at $T = 30$ captures at most $9.6\%$ of it, so the observed gaps lie far below what any procedure exploiting the test set could achieve under this budget.

\FloatBarrier
\needspace{0.6\textheight}
\FloatBarrier
\section{MLP-Giant$_{T{=}100}$: Adaptive Overfitting Under Increased Tuning and Data Budget}
\label{app:robustness}

\begin{figure}[!htbp]
  \centering
  \begin{minipage}[t]{0.43\linewidth}
    \centering
    \includegraphics[width=\linewidth]{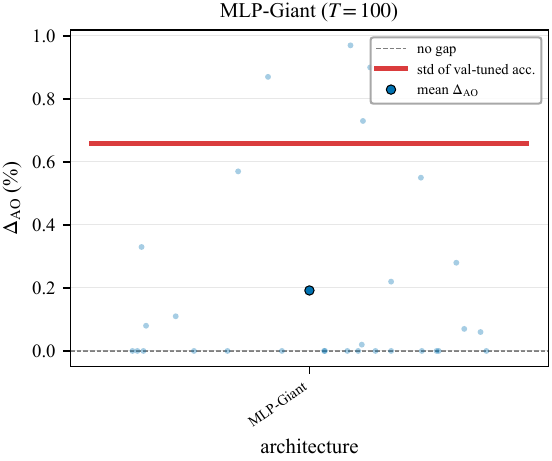}
  \end{minipage}\hspace{0.5em}\begin{minipage}[t]{0.43\linewidth}
    \centering
    \includegraphics[width=\linewidth]{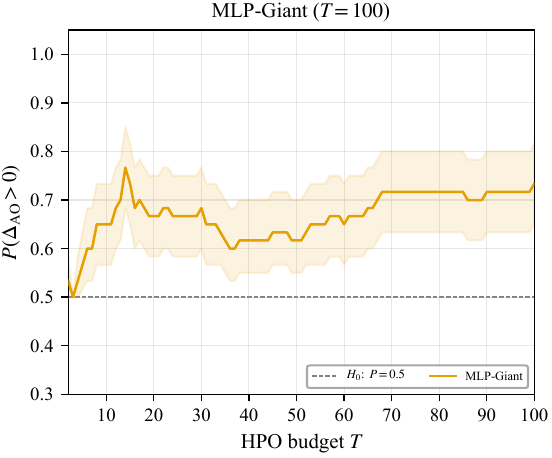}
  \end{minipage}
  \caption{MLP-Giant rerun with $T = 100$ trials per run on larger $10{,}000 / 10{,}000 / 10{,}000$ MNIST-1D splits. \textbf{Left:} per-run $\Delta_{\mathrm{AO}}$ across the $K = 30$ paired runs; the blue dot is the mean and the red horizontal bar marks the standard deviation of val-tuned accuracy. \textbf{Right:} $\hat{P}(\Delta_{\mathrm{AO}} > 0)$ as a function of hyperparameter tuning budget $T$, with the $95\,\%$ percentile-bootstrap CI; the dashed grey reference at $0.5$ is the significance threshold. The mean gap stays small relative to the val-tuned noise and $\hat{P}(\Delta_{\mathrm{AO}} > 0)$ remains significant ($\text{CI}_{\text{lower}} > 0.5$) across the full budget range.}
  \label{fig:robustness_t100}
\end{figure}

\FloatBarrier
\section{Adaptive Overfitting Transfer to CIFAR-10.1}
\label{app:cifar101}

We re-evaluated the selected CIFAR-10 configurations on CIFAR-10.1~\citep{recht2019imagenet_generalize}, a $2{,}000$-image replication of CIFAR-10 drawn from the same source distribution. On this independent set the val-tuned and test-tuned configurations score within run-to-run noise of each other, so the gap measured on the bootstrap fold does not carry over. Tuning directly on CIFAR-10.1 beats both by roughly $0.4\%$, showing what a procedure with access to it could gain (Appendix~\ref{app:cifar101}).

\begin{table}[!htbp]
  \centering
  \footnotesize
  \setlength{\tabcolsep}{2pt}
  \renewcommand{\arraystretch}{0.85}
  \begin{tabular}{l ccc}
    \toprule
    Model & Val-tuned (\%) & Test-tuned (\%) & CIFAR-10.1-tuned (\%) \\
          & on CIFAR-10.1 & on CIFAR-10.1 & on CIFAR-10.1 \\
    \midrule
    \texttt{shake\_shake\_32d} & $88.79$ & $88.82$ & $89.11$ \\
    \texttt{resnet9}           & $86.95$ & $86.93$ & $87.41$ \\
    \texttt{mobilenetv2}       & $87.11$ & $87.06$ & $87.29$ \\
    \texttt{resnet\_basic\_56} & $86.07$ & $86.11$ & $86.39$ \\
    \texttt{resnet\_basic\_44} & $85.95$ & $85.78$ & $86.26$ \\
    \texttt{vgg\_15\_BN\_64}   & $85.40$ & $85.34$ & $85.86$ \\
    \texttt{resnet\_basic\_32} & $85.23$ & $85.11$ & $85.69$ \\
    \texttt{resnet\_basic\_20} & $83.81$ & $83.81$ & $84.36$ \\
    \midrule
    \textbf{Pooled} & $\mathbf{86.16}$ & $\mathbf{86.12}$ & $\mathbf{86.55}$ \\
    \bottomrule
  \end{tabular}
\caption{Per-architecture mean CIFAR-10.1 accuracy under three selection rules. \textbf{Val-tuned} and \textbf{Test-tuned} pick the trial with the highest accuracy on the bootstrap validation or test fold, respectively, and the picked configuration is then re-evaluated on CIFAR-10.1; \textbf{CIFAR-10.1-tuned} picks the trial with the highest CIFAR-10.1 accuracy directly. ``Pooled'' is the mean across the eight architectures. The val-tuned and test-tuned columns are indistinguishable on CIFAR-10.1 (pooled gap $-0.04\%$, within run-to-run noise), consistent with the small $\Delta_{\mathrm{AO}} \approx 0.12\%$ already observed on the bootstrap test fold. CIFAR-10.1-tuned wins by $+0.43\%$ pooled --- a small but consistent margin that the bootstrap-test-tuned configuration does not capture.}
  \label{tab:cifar101_results}
\end{table}

\FloatBarrier
\section{Data Splitting}
\label{app:data-splitting}

We use the same splitting scheme on all five benchmarks. We start from a single source pool: $4{,}000$ examples for MNIST-1D, the $60{,}000$-example pool that combines CIFAR-10's training and test sets, and for the GLUE tasks the concatenation of the official train and validation splits ($2{,}767$ for RTE, $4{,}076$ for MRPC and $9{,}594$ for CoLA). For each paired run we draw the train set by sampling from the pool with replacement, taking as many samples as the pool size. Examples that were never picked end up in the leftover set, which on average is about $36.8\%$ of the pool. We shuffle this leftover set and split it in half: one half is the validation set and the other is the test set.

For MNIST-1D this gives roughly $4{,}000$ training, $736$ validation and $736$ test examples per run, and for CIFAR-10 roughly $60{,}000$ training, $11{,}040$ validation and $11{,}040$ test examples. For the GLUE tasks it gives roughly $509$ validation and $509$ test examples on RTE, $750$ and $750$ on MRPC, and $1{,}763$ and $1{,}763$ on CoLA. The validation and test sets never overlap, and they never share an example with the training set. The split is drawn once at the start of a run and stays the same across all $T$ trials, so validation-tuning and test-tuning pick from the same pool of trained models. A new split is drawn for each of the $K = 30$ paired runs.

We use the pooled train and validation splits for the GLUE tasks because the official GLUE test splits are unlabeled and withheld on the leaderboard server, so the validation split is the de facto test set in community practice.

The MLP-Giant rerun in Appendix~\ref{app:robustness} uses a different scheme. It draws three disjoint sets of $10{,}000$ examples each (train, validation, test) from a larger $70{,}000$-example MNIST-1D set, without replacement.

\FloatBarrier
\FloatBarrier
\section{Hyperparameter Range Validity}
\label{app:hp-validity}

\begin{figure}[!htbp]
  \centering
  \includegraphics[width=0.50\linewidth]{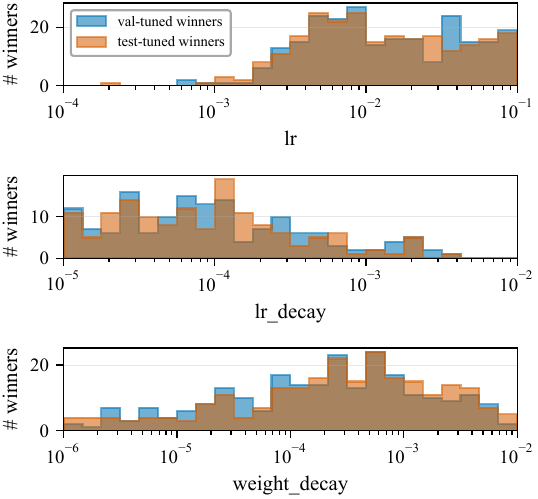}
    \caption{MNIST-1D: empirical distribution of winning hyperparameter values for the validation-tuned and test-tuned selection rules. $30.8\%$ of trials had $\texttt{lr\_decay} = 0$ and are excluded from the $\texttt{lr\_decay}$ panel. All three HPs have their modes well inside the bounds with no edge clustering.}
  \label{fig:hp-validity-mnist1d}
\end{figure}

\begin{figure}[!htbp]
  \centering
  \includegraphics[width=0.50\linewidth]{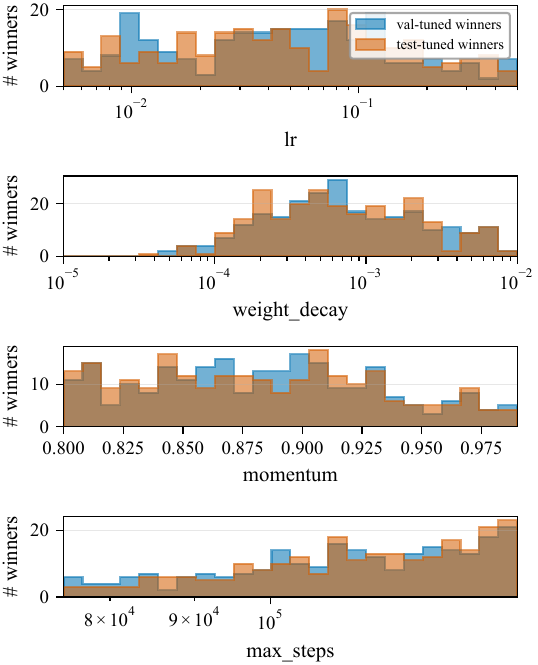}
\caption{CIFAR-10: empirical distribution of winning hyperparameter values for the val-tuned and test-tuned selection rules. $\texttt{lr}$, $\texttt{weight\_decay}$, and $\texttt{momentum}$ have their modes well inside the bounds; only $\texttt{max\_steps}$ clusters at the upper bound of $141{,}000$, suggesting the optimum lies above this cap. Extending it would raise absolute accuracies but is unlikely to change $\Delta_{\mathrm{AO}}$, since both selection rules benefit equally from longer training.}
  \label{fig:hp-validity-cifar10}
\end{figure}

\begin{figure}[!htbp]
  \centering
  \includegraphics[width=0.50\linewidth]{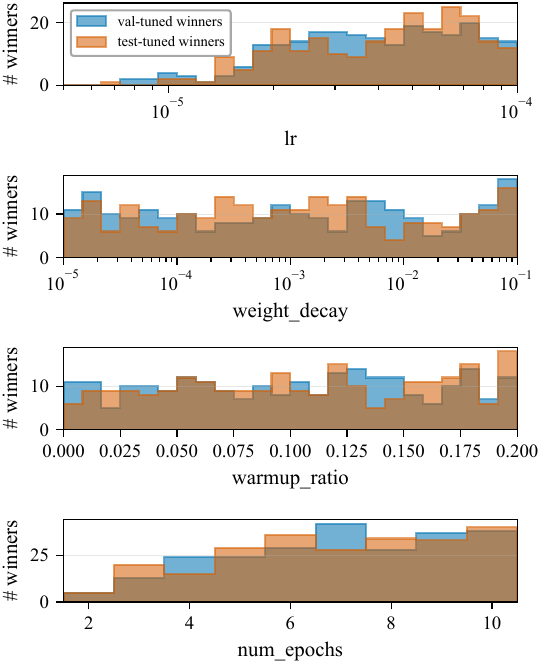}
  \caption{RTE: empirical distribution of winning hyperparameter values for the val-tuned and test-tuned selection rules, over the $240$ winners of each rule ($8$ architectures $\times$ $K = 30$ runs). $\texttt{num\_epochs}$ is binned on the integers $2$--$10$.}
  \label{fig:hp-validity-rte}
\end{figure}

\begin{figure}[!htbp]
  \centering
  \includegraphics[width=0.50\linewidth]{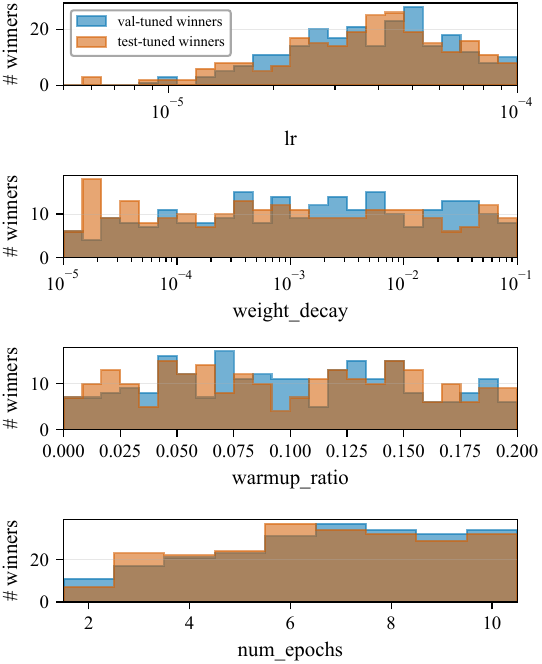}
  \caption{MRPC: empirical distribution of winning hyperparameter values for the val-tuned and test-tuned selection rules, over the $240$ winners of each rule.}
  \label{fig:hp-validity-mrpc}
\end{figure}

\begin{figure}[!htbp]
  \centering
  \includegraphics[width=0.50\linewidth]{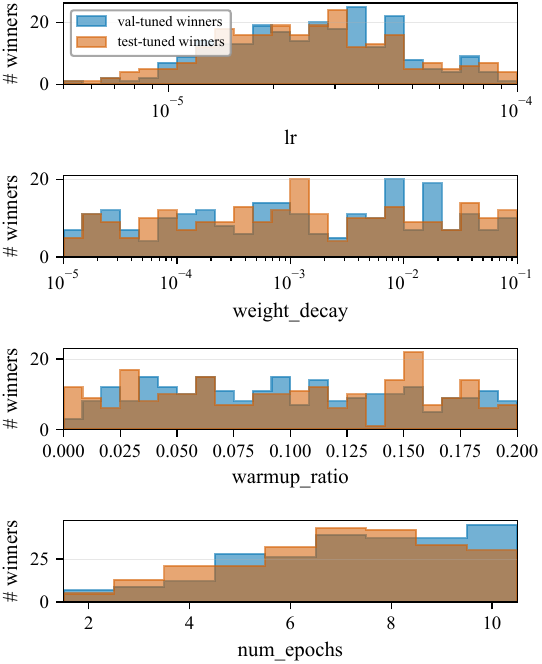}
  \caption{CoLA: empirical distribution of winning hyperparameter values for the val-tuned and test-tuned selection rules, over the $240$ winners of each rule. As on the other two GLUE tasks, $\texttt{num\_epochs}$ rises towards the upper bound of the range.}
  \label{fig:hp-validity-cola}
\end{figure}

The empirical winner distributions in Figures~\ref{fig:hp-validity-mnist1d} and~\ref{fig:hp-validity-cifar10} suggest the search ranges are well-shaped: $\texttt{lr}$, $\texttt{weight\_decay}$, and $\texttt{momentum}$ have their modes inside the bounds with no edge clustering. The one exception is CIFAR-10's $\texttt{max\_steps}$, whose winners pile up at the upper bound of $141{,}000$; extending the range would benefit both selection rules equally and is unlikely to qualitatively change $\Delta_{\mathrm{AO}}$. For the GLUE tasks the ranges were instead set a priori to bracket the published fine-tuning grids of all eight model families (Table~\ref{tab:hpo-spaces-glue}), with one deliberate exception: ELECTRA-small's published $3 \times 10^{-4}$ recipe lies above our $10^{-4}$ learning-rate cap, which is kept below the divergence region of the base-size models. The winner distributions in Figures~\ref{fig:hp-validity-rte}--\ref{fig:hp-validity-cola} confirm those ranges empirically: $\texttt{lr}$ peaks around $3$--$5 \times 10^{-5}$, where the published recipes sit, while $\texttt{weight\_decay}$ and $\texttt{warmup\_ratio}$ are close to flat, indicating that neither strongly determines which configuration wins. As with CIFAR-10's $\texttt{max\_steps}$, only $\texttt{num\_epochs}$ clusters at its upper bound of $10$, and extending it would again benefit both selection rules equally.

\FloatBarrier
\section{Hyperparameter Importance}
\label{app:fanova}

\begin{table}[!htbp]
\centering
\caption{fANOVA main-effect importances for $\Delta_{\mathrm{AO}}$.  Each entry is the mean over the eight architectures of a benchmark, followed by its share of the benchmark's total listed importance.}
\label{tab:fanova}
\small
\begin{tabular}{lccccc}
\toprule
 & MNIST-1D & CIFAR-10 & RTE & MRPC & CoLA \\
\midrule
learning rate      & 0.142 (30\%) & 0.146 (30\%) & 0.106 (24\%) & 0.126 (28\%) & 0.141 (32\%) \\
weight decay       & 0.105 (23\%) & 0.107 (22\%) & 0.094 (21\%) & 0.089 (20\%) & 0.101 (23\%) \\
lr decay rate      & 0.090 (19\%) & ---          & ---          & ---          & ---          \\
momentum           & ---          & 0.075 (15\%) & ---          & ---          & ---          \\
training steps     & ---          & 0.077 (16\%) & ---          & ---          & ---          \\
warmup ratio       & ---          & ---          & 0.107 (24\%) & 0.096 (21\%) & 0.096 (22\%) \\
epochs             & ---          & ---          & 0.035 (8\%)  & 0.040 (9\%)  & 0.028 (6\%)  \\
seed               & 0.129 (28\%) & 0.080 (16\%) & 0.104 (23\%) & 0.097 (22\%) & 0.078 (18\%) \\
\bottomrule
\end{tabular}
\end{table}

Does the gap come from a particular hyperparameter? We fit one fANOVA surrogate per architecture, with $\Delta_{\mathrm{AO}}$ as the target and that benchmark's search space as the domain, and average the main-effect importances over the eight architectures (Table~\ref{tab:fanova}). The initialisation seed is one of the tuned hyperparameters (Appendix~\ref{app:hpo-spaces}), so it appears alongside the others.

No single hyperparameter dominates. The learning rate is the largest effect on every benchmark, but only at $24$--$32\%$ of the total, with weight decay next at $20$--$23\%$ and the benchmark-specific hyperparameters below that. The seed contributes $16$--$28\%$, so the remaining hyperparameters together account for $72$--$84\%$: tuning on the test set is not simply picking a lucky initialisation. Importance is spread fairly evenly across the whole search space, which is what we expect if the gap comes from noise that tuning selects on rather than from any particular region of it.

\end{document}